\documentclass[letterpaper]{article} 
\usepackage[preprint]{aaai2027}  
\usepackage[hyphens]{url}  
\usepackage{graphicx} 
\usepackage{natbib}  
\usepackage{caption} 
\usepackage{algorithm}
\usepackage{algpseudocode}

\usepackage{newfloat}
\usepackage{listings}
\DeclareCaptionStyle{ruled}{labelfont=normalfont,labelsep=colon,strut=off} 
\floatstyle{ruled}
\newfloat{listing}{tb}{lst}{}
\floatname{listing}{Listing}

\usepackage{booktabs}
\usepackage{multirow}
\usepackage{amsmath}
\usepackage{amssymb}

\newcommand{\squishlist}{
 \begin{list}{$\bullet$}
  { \setlength{\itemsep}{0pt}
     \setlength{\parsep}{3pt}
     \setlength{\topsep}{3pt}
     \setlength{\partopsep}{0pt}
     \setlength{\leftmargin}{1.5em}
     \setlength{\labelwidth}{1em}
     \setlength{\labelsep}{0.5em} } }

\newcommand{\squishend}{
  \end{list}  }

\definecolor{Georgia}{RGB}{132, 2, 22}

\title{Semantic-aware Decoding for Diverse Text-to-SQL under Schema Ambiguity}
\author{
    Yuxuan Jiang\textsuperscript{\rm 1, \rm 2}, Anna Mitsopoulou\textsuperscript{\rm 1}, Apostolos Koukouvinis\textsuperscript{\rm 1}, Yannis Velegrakis\textsuperscript{\rm 2}, Georgia Koutrika\textsuperscript{\rm 1}\corresponding\\
}
\affiliations{
    \textsuperscript{\rm 1}ATHENA Research Center, Greece\\
    \textsuperscript{\rm 2}Utrecht University, Netherlands\\
    yjiangei@athenarc.gr, georgia@athenarc.gr
}

\begin{document}

\maketitle

\begin{abstract}
Real-world text-to-SQL systems often face schema ambiguity, where a question can map to multiple plausible schema elements. To support downstream disambiguation, SQL generation systems should include diverse interpretations among their top-k predictions. However, large language models (LLMs) typically favor a single interpretation, producing highly skewed outputs. We propose Semantic-aware Logit Adjustment (SLA), a training-free decoding method that promotes diverse SQL generation by adjusting token logits based on semantic similarity between schema elements. Unlike prior approaches that rely on reasoning-intensive prompting or task-specific fine-tuning, SLA is efficient, model-agnostic, and easy to deploy. Experiments show that SLA consistently improves interpretation coverage over standard decoding and remains competitive with existing diversity-oriented methods at substantially lower inference cost.
\end{abstract}


\section{Introduction}

Large language models have shown remarkable performance in many tasks. Among them, text-to-SQL plays an important role in transforming user natural language questions into executable SQL queries. Over recent years, many works have made significant progress on benchmarks such as SPIDER\cite{spider} and BIRD\cite{bird}, indicating the potential of LLMs as an assistant for SQL writing.

\begin{figure*}[t]
    \centering
    \includegraphics[width=0.85\linewidth]{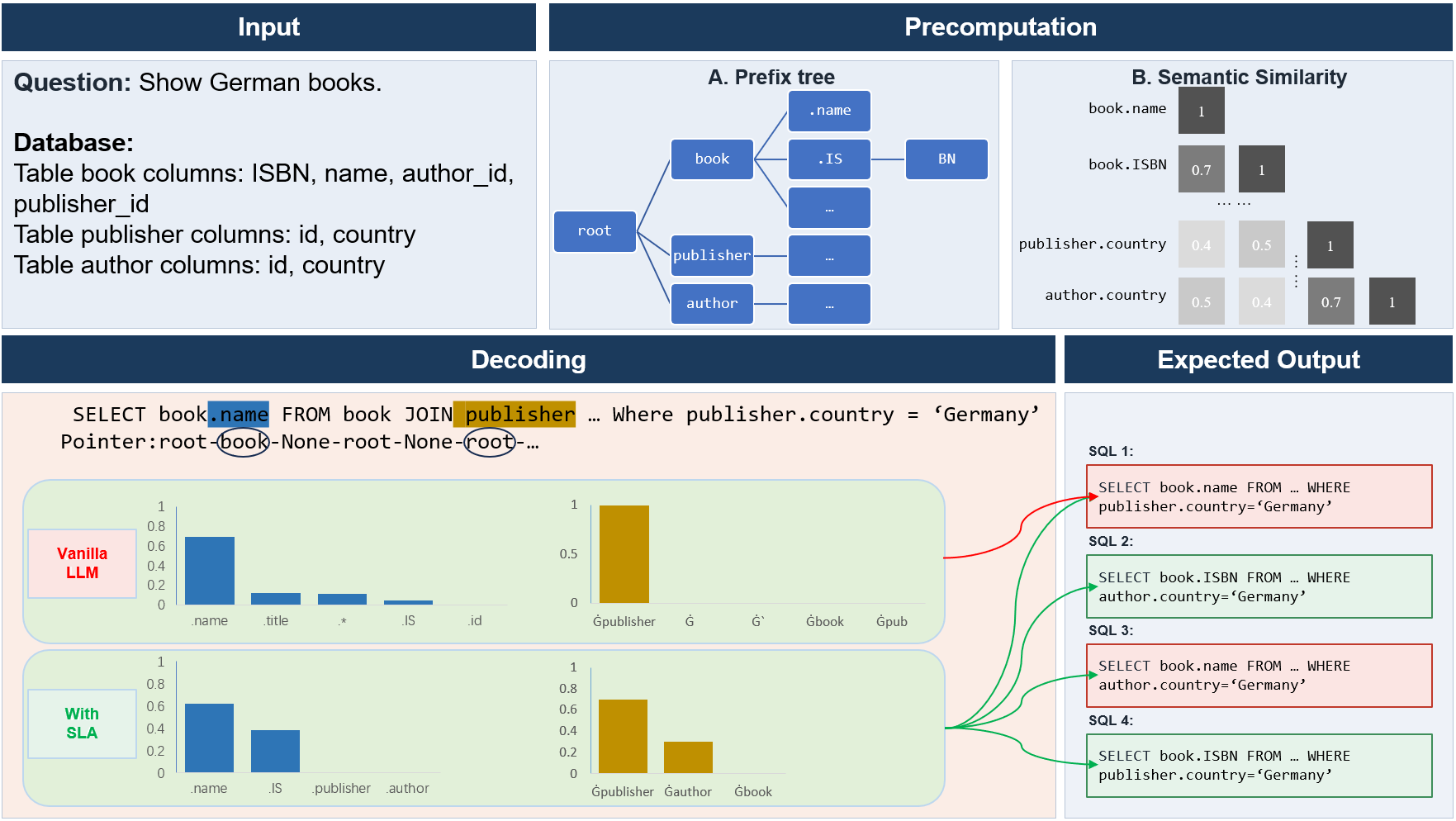}
    \caption{An illustration example of our framework. Given a user question over a database, we first precompute the prefix tree and pairwise semantic similarities of all schema elements, and then adjust the output distribution based on the pointer and similarity. As a result, our method facilitates the model to explore more potential schema interpretations.}
    \label{fig:overview_example}
\end{figure*}

In practice, however, the questions people actually ask are rarely so well posed. A user question can be ambiguous, and several different but equally valid SQL queries could answer it\cite{floratou2024nl2sql}, whereas most existing benchmarks sidestep this by assuming that each question is clear and corresponds to a single ground truth. The source of ambiguity can roughly be classified into three types: database schema, database value and natural language. Among them, schema ambiguity \cite{logicalbeam, know_what_idk, floratou2024nl2sql} refers to the scenario where an entity in the user question can refer to different database schema elements (tables or columns). It is a major source of failure in real-world applications\cite{know_what_idk}, because the database can contain similar concepts, users are unfamiliar with the database, and natural language expressions often omit crucial expressions. For example in Fig.
\ref{fig:overview_example}, it is unclear whether the user wants the name or the ISBN of the books, and whether the books should be filtered by German writers or by German publishers. Since the system cannot resolve the ambiguity in such a question on the user's behalf, the practical requirement is that every plausible reading survives: all valid interpretations should appear among the top-$k$ predictions, so that they can be presented to the user or passed to a disambiguation module. We refer to this property as interpretation coverage.

The most natural way to obtain such a set is diverse decoding, such as beam search and random sampling, which is efficient and requires no additional training. Previous work has shown, however, that these methods are ineffective at generating semantically different SQL queries, mainly due to highly skewed logit distributions\cite{logicalbeam}: at the very step where the interpretations diverge, a single token dominates the distribution and the alternative is buried far below. Existing works resort to alternative paradigms, like reasoning-intensive prompting\cite{ambisql, clear} or task-specific fine-tuning\cite{DFPL, IntentRL}. While effective, they have two main limitations: (1) they introduce high computational cost due to test-time scaling or the training procedure; (2) they have weak applicability to a broader range of base models, as they assume reasoning-capable or task-adapted LLMs. We therefore ask whether the skewed distribution can instead be repaired during decoding, which would retain the efficiency, the training-free nature and the broad applicability that made diverse decoding attractive in the first place: \textit{can we facilitate interpretation coverage by improving the decoding itself?}

\begin{figure}[t]
    \centering
    \includegraphics[width=0.7\linewidth]{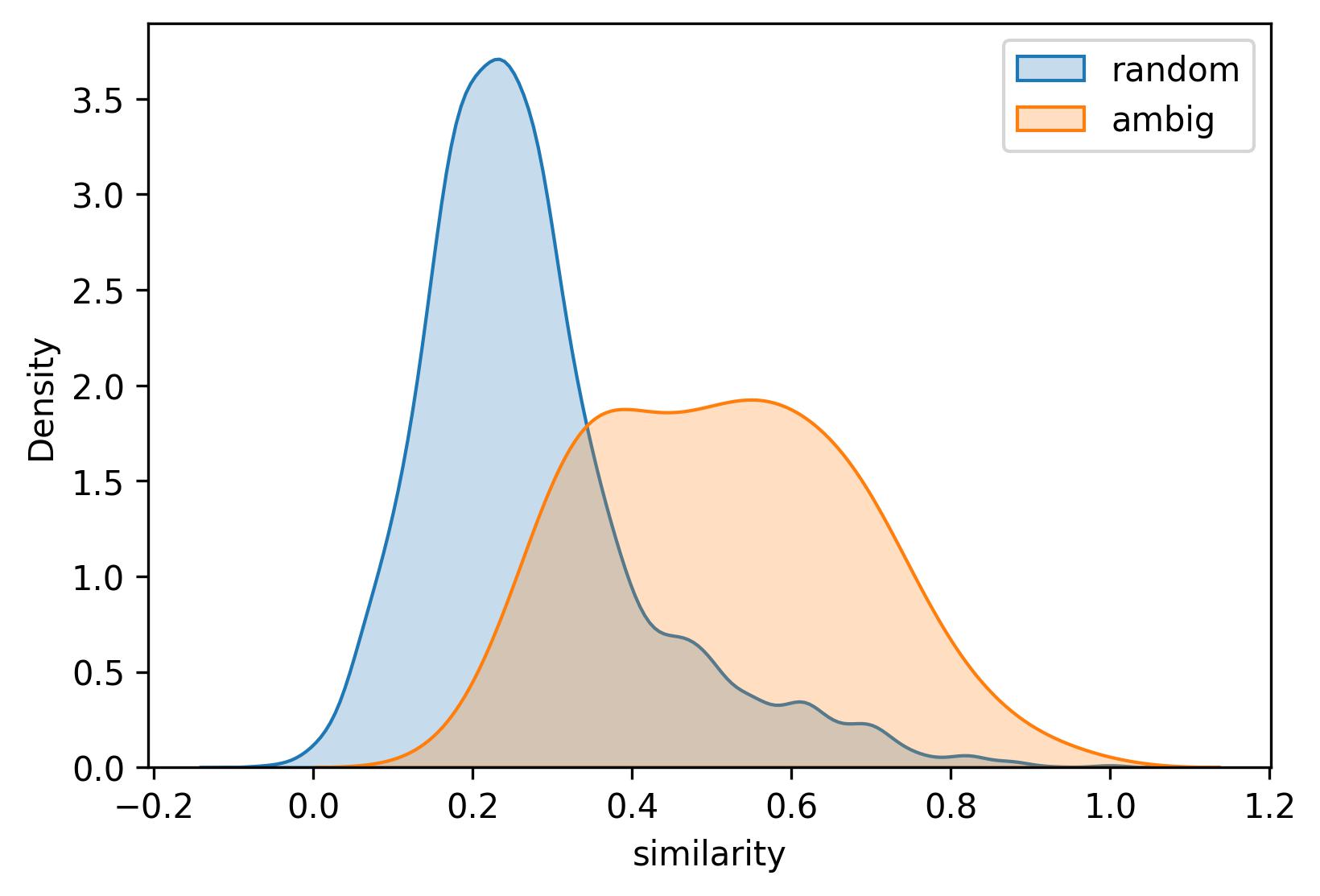}
    \caption{Similarity distribution. We randomly sample 80 examples from AmbiQT and Ambrosia, and compute the cosine similarity of embeddings between random schema element pairs and schema element pairs that induce ambiguity.}
    \label{fig:sim_distribution}
\end{figure}

Our answer starts from a simple observation: schema ambiguity typically arises among schema elements that describe closely related concepts, such as the author and the publisher, which both contribute to the release of a book. Such conceptual relatedness can be quantified by semantic similarity between schema element embeddings. As shown in Fig.~\ref{fig:sim_distribution}, schema elements that induce ambiguity show higher semantic similarities than random schema pairs on average. Therefore, semantic similarity provides an effective prior for identifying alternative schema interpretations. 
We exploit it directly at decoding time: whenever the model is about to write a schema element, we raise the scores of the elements that are semantically closest to the one it currently favors, so that plausible alternatives rise out of the tail of the distribution and survive in the output set.

We evaluate our method on two benchmarks that contain schema ambiguity, across base models of different families and sizes. Our method significantly improves the interpretation coverage over conventional decoding methods and achieves performance comparable to methods that rely on fine-tuning or test-time scaling, while adding only about 10\% latency over vanilla beam search.

\section{Related work}

Existing work handles ambiguous questions by generating a set of SQL queries, which are either presented to users directly\cite{odin, DFPL} or used for downstream disambiguation\cite{pleasqlarify, EIG_ambi, clear}. Both settings rely on interpretation coverage.

For general ambiguity, DFPL\cite{DFPL} prompts LLMs to produce multiple natural language interpretations, augments them with a fine-tuned model and maps each to a SQL query, IntentRL\cite{IntentRL} formulates interpretation generation as a reinforcement learning problem to explicitly encourage diversity, and CLEAR\cite{clear} exploits the reasoning ability of models with few-shot examples to map entities in the question to schema elements; none of them focuses on schema ambiguity specifically. Targeting schema ambiguity specifically, LogicalBeam\cite{logicalbeam} trains two models to first generate SQL templates and then fill them with schema elements, and ODIN\cite{odin} masks the schema elements used in previous predictions to enforce the exploration of alternatives.

All of these methods intervene on the input or on the model weights and leave the decoding algorithm itself unchanged, paying for diversity with extra prompts, extra generation passes or extra training. In contrast, we directly adjust the token distribution during decoding, which yields diverse SQL queries within a single parallel decoding pass.



\section{Problem definition}
\paragraph{Schema ambiguity} We first provide a definition of schema ambiguity in the context of text-to-SQL. Following previous work\cite{floratou2024nl2sql}, a question is ambiguous if multiple non-equivalent and acceptable SQL queries can answer it. When the natural language itself is clear, schema ambiguity occurs when there is more than one way to map the question to database schema.

\paragraph{Problem formulation} Given a database schema \(D\), consisting of \(N\) schema elements \(C=[c_1, c_2, ..., c_N]\), and a user question \(Q\) with schema ambiguity, there exist multiple non-equivalent but valid queries \(\{y_i\}\) that answer it. Our aim is to utilize a text-to-SQL model \(P_\theta\) (usually an LLM) to generate a set of outputs \(\{\hat{y}_i\}\), formally 
\[
\hat{y}_i \sim P_\theta(y|Q, D)
\]
such that the prediction set captures all possible interpretations, i.e., \(\{y_i\} \subseteq \{\hat{y}_i\}\).

\section{Method}

\subsection{Method overview}

Consider a database with tables book, author and publisher, and the question "Show German books" (Fig.~\ref{fig:overview_example}). Two interpretations are valid: books written by German authors, and books released by German publishers. The model has to choose between them at one decoding step, the step that generates the table name after JOIN. At that step, the logit distribution is highly skewed: one token dominates the probability distribution, while the other one is ranked much below. The second interpretation is lost because this skewed distribution hides it, so beam search or sampling will never reach this interpretation.

Our key observation is that alternative interpretations of an ambiguous query come from schema elements that describe closely related concepts. Author and publisher both relate to the production of a book; name and ISBN both identify a book. We quantify this by embedding each schema element with an embedding model and calculating their pairwise cosine similarities. Schema elements that cause ambiguity have high similarity, while unrelated pairs usually have low similarity (Fig.~\ref{fig:sim_distribution}). Although semantic similarity alone does not imply ambiguity, ambiguous schema elements almost always exhibit substantial semantic overlap.

This leads to our method. Whenever the model is generating a schema element, we add to the logit of each candidate token a boost that depends on the schema similarities. For example, after generating \texttt{book}, the top-1 prediction for the next token is \texttt{.name}. We boost \texttt{.IS} because it leads to \texttt{book.ISBN}, which is semantically similar to \texttt{book.name}, while unrelated tokens receive little or no boost. In the following sections, we first describe how we decide whether to modify the logits at the current step (Section~\ref{sec: when_to_modify}), and then present our algorithm for computing the boost (Section~\ref{sec:SLA}).


\subsection{SQL generation tracking}
\label{sec: when_to_modify}


We first need to decide two things: when to modify the logits, and for which tokens. A SQL query consists of keywords, database values and schema elements. We intervene only while a schema element is being generated and only on tokens that are valid continuations of the schema element generated so far. Two problems make this non-trivial. First, the model generates tokens one by one, and nothing in the token itself tells us whether it belongs to a schema element. Second, a schema element often consists of several tokens, for example \texttt{ISBN} may be split into \texttt{IS} and \texttt{BN}, so that for a prefix token we do not know which schema element the model is generating. What we need is therefore a mechanism that, at every step, tells us whether the model is inside a schema element and, if so, which tokens may follow.

To achieve that we create a prefix tree\cite{TTD-SQL} with all the schema elements. More specifically, in an \textit{offline stage}, we tokenize all schema elements, i.e., table.column for the columns, and insert them into a token-level prefix tree, in which every path from the root to a leaf corresponds to a schema element. Then, \textit{during decoding}, maintain a pointer over the tree, which records the schema element generation point at each decoding step. The pointer has two possible states: (1) None, when the model is not generating any schema element or the model starts to generate a new clause (2) prefix tree node, while a schema element is being generated. In each step we update the pointer as follows: if the pointer is None, we check if the next top-1 token matches a child of the root node of a prefix tree and we set the pointer to root if this is the case. If the pointer is at root or at an inner node of the prefix tree, we advance the pointer to the child matching the last generated token. When the pointer has reached a leaf node, we set the pointer to None, if the next-token is not a child of the root, otherwise it is set to root.

In the example, the pointer is \texttt{None} while the model generates \texttt{SELECT}, so the logits are unchanged. When the model generates \texttt{book}, the pointer advances to the internal node reached by \texttt{root-book}, indicating that the model is inside a schema element. Only the node's children, such as \texttt{.name} and \texttt{.IS}, are valid continuations; \texttt{.title} is treated as hallucinated and suppressed by logit adjustment. Once the element is complete, the pointer reaches a leaf, returns to \texttt{None}, and decoding continues normally.

Logits are adjusted only when the pointer is at the root node or at an inner node state and the tokens to be adjusted become only the children of the pointed node.

\begin{figure}[h]
    \centering
    \includegraphics[width=0.7\linewidth]{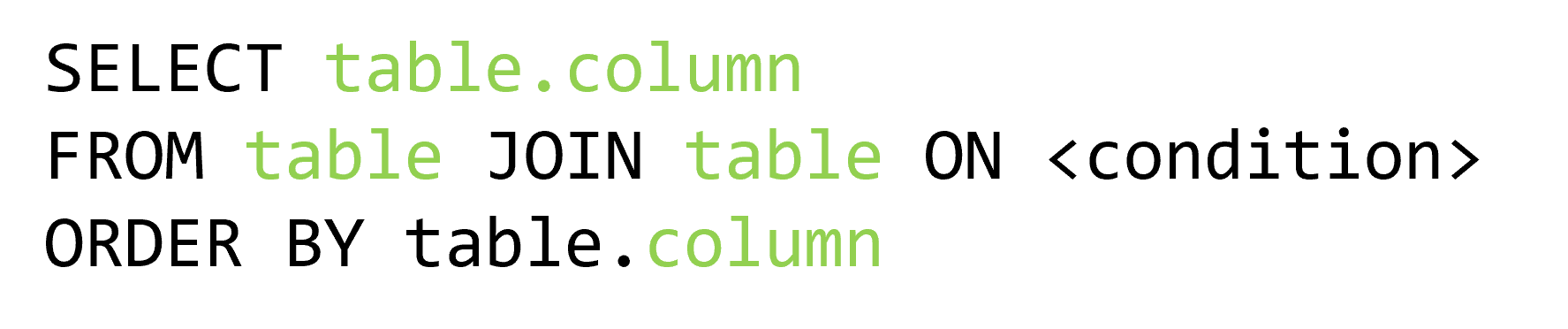}
    \caption{Example SQL query template. The ORDER BY clause can be replaced by other clauses such as WHERE, GROUP BY. Positions where the logits are modified are highlighted in green.}
    \label{fig:pos_to_modify}
\end{figure}

Finally, we enable the mechanism only where schema alternatives are meaningful: table names in \texttt{SELECT}, \texttt{FROM}, and \texttt{JOIN}, and column names in \texttt{SELECT}, \texttt{WHERE}, \texttt{GROUP BY}, \texttt{HAVING}, and \texttt{ORDER BY}. We exclude columns inside \texttt{COUNT} to avoid unnecessary randomness, and the \texttt{ON} clause because its columns are fixed by primary--foreign key relationships; alternatives would invalidate the join condition. Fig.~\ref{fig:pos_to_modify} illustrates these positions.

\subsection{Semantic-aware logit adjustment (SLA)}
\label{sec:SLA}

\begin{algorithm}[tb]
\caption{SLA at one decoding step}
\label{alg:logitsprocessor}
\begin{algorithmic}[1]
\State {\bfseries Precomputed once per database:} prefix tree \(T\), similarity matrix \(S\)
\State {\bfseries Input:} tokens generated so far \(o\), logits \(z\), pointer \(P\), strength \(\alpha\)
\State {\bfseries Output:} adjusted logits \(z\)
\State \(P \gets\) \texttt{UpdatePointer}(\(P\), \(o\), \(z\), \(T\)) \Comment{\textit{how far we are into a schema element}}
\If{\textbf{not} \texttt{InTargetClause}(\(o\)) \textbf{or} \(P\) is \texttt{None} \textbf{or} \(P\) is a leaf}
    \State \Return \(z\) \Comment{\textit{keyword, value, finished element, or skipped clause}}
\EndIf
\State \(A \gets\) \texttt{Children}(\(P\)) \Comment{\textit{allowed next tokens}}
\State \(z[d] \gets -\infty\) \textbf{for all} \(d \notin A\)
\State \(q \gets \arg\max_{d \in A} z[d]\) \Comment{\textit{the model's own preference}}
\State \(C_q \gets\) \texttt{Elements}(\(q\)) \Comment{\textit{schema elements below node \(q\)}}
\State \(\lambda \gets \alpha(z_q-z_{q'})\) \Comment{\textit{keeps \(\arg\max\) unchanged}}
\For{\(d \in A\), \(d \neq q\)}
    \State \(C_d \gets\) \texttt{Elements}(\(d\))
    \State \(z[d] \gets z[d] + \lambda \cdot \mathrm{sim}(C_q, C_d)\) 
\EndFor
\State \Return \(z\)
\end{algorithmic}
\end{algorithm}

We proceed in four steps, pointing to the lines of Algorithm~\ref{alg:logitsprocessor}.

\paragraph{Step 1: Similarity between schema elements.}
Each schema element is represented by a short text, the table name for a table and \texttt{table.column} for a column. We encode it with an off-the-shelf sentence embedding model, and store all pairwise cosine similarities in \(S \in \mathbb{R}^{N\times N}\). This happens once per database, before decoding (line~1). Raw cosine values are not comparable across embedding models: one may place most schema pairs in \([0.6,0.9]\), another in \([0.0,0.4]\). Since our boost is proportional to them, a fixed hyper-parameter would act differently for each model. Only the relative order of the pairs matters to us, so we apply histogram equalization and replace each similarity by the fraction of pairs scoring below it. This spreads the values of any model uniformly over \([0,1]\), so one setting of the strength parameter transfers across models.

\paragraph{Step 2: From element similarity to token similarity.}
\(S\) relates elements, but decoding boosts tokens, and a token is usually only a prefix: \texttt{.IS} is not an element but the start of \texttt{book.ISBN}, and the node reached by \texttt{author} starts \texttt{author.name}, \texttt{author.country} and \texttt{author.id}. The prefix tree links the two. Every candidate token \(d\) is a child of the pointer, and the leaves of its subtree are exactly the elements still reachable through \(d\); we write \(C_d\) for that set and \(C_q\) for the set below the node \(q\) the model currently prefers (lines~11 and~14). Node similarity is therefore a similarity between two sets of elements.

A plain average over all pairs in \(C_q \times C_d\) does not work. Even when two nodes stand for related concepts, only a few pairs of elements really match. All the other pairs are unrelated, and they drag the average down. The more elements the two nodes cover, the more such pairs there are, so a table with many columns would look dissimilar to everything, simply because it has many columns. What we really ask is whether every element reachable through \(d\) has a close match under \(q\). We therefore pair each element of \(C_d\) with its most similar element in \(C_q\), and average over these pairs:
\begin{equation}
\label{eq:sim_calculation}
    \mathrm{sim}(C_q, C_d)=\frac{1}{|C_d|}\sum_{j\in C_d}\max_{i\in C_q}S[i,j]
\end{equation}
The same match-then-average pattern appears in BERTScore~\cite{bertscore} and in the late interaction of ColBERT~\cite{colbert}. We compare both choices in our ablation study.

\paragraph{Step 3: Where the boost is added.}
The model produces logits \(z \in \mathbb{R}^V\) with \(p=\mathrm{softmax}(z)\). Editing \(p\) is tempting, but the softmax hides what we need: in a skewed distribution every token below the top has probability close to zero, and a plausible alternative becomes indistinguishable from an impossible one, although their logits still differ clearly. We therefore adjust the logits, \(z' = z + b\). In general \(b\) depends on the whole distribution, \(b(S,z)\); we simplify it to \(b(S,q)\) by scoring every candidate against the token \(q\) the model itself ranks first (line~10). This is safe because the element ranked first is almost always in one of the valid queries or close to one, and it reduces the cost from \(|A|^2\) to \(|A|\) similarities, where \(A\) is the set of allowed tokens. We also never operate over the full vocabulary of the model: the children of the pointer are the only possible continuations, so \(A\) is exactly that set of children and everything outside it is masked to \(-\infty\) (lines~8--9).

\paragraph{Step 4: How strong the boost is.}
For a candidate \(d\) we set \(b_q(d)=\lambda\cdot\mathrm{sim}(C_q,C_d)\) (line~15), where \(\lambda\) scales the similarity, which lies in \([0,1]\),  up to the scale of the logits. \(\lambda\) should grow when the model is overconfident, since that is when the alternative is hidden, but deriving it from the entropy is unstable: a large \(\lambda\) may overturn the top-1 and change the query the model was about to write. Instead, we impose \(\arg\max(z')=\arg\max(z)\), so that SLA re-ranks alternatives but never overrules the first choice. Writing \(q\) for the allowed token with the largest logit and \(q'\) for the runner-up, we set \(\lambda=\alpha\,(z_q-z_{q'})\) with \(0<\alpha<1\), a fraction of the gap between the two (line~12). As \(\mathrm{sim}\le 1\), the largest possible boost is \(\alpha\) times that gap, strictly below the gap itself, so no candidate can overtake \(q\).

\subsection{SchemaBeam}
\label{sec:app}
Because our algorithm operates by modifying logits, it is compatible with both deterministic and stochastic decoding methods, such as beam search and random sampling. In practice, beam search provides better interpretation coverage, so we introduce SchemaBeam, which integrates SLA into standard beam search. Vanilla beam search, however, often produces unhelpful token-level diversity, including formatting variants (e.g., trailing semicolons or code fences) and logically equivalent queries that differ only in join order or other syntactic details. SchemaBeam reduces this redundancy by masking related tokens during decoding and deduplicating outputs syntactically using their ASTs.


\section{Experiments}

\subsection{Setup}

\paragraph{Datasets} 

We select two benchmarks that contain one or more ambiguity categories related to schema ambiguity:
\squishlist
    \item \textit{AmbiQT}\cite{logicalbeam} This dataset is modified from Spider \cite{spider} and contains fine-grained schema ambiguity types. For each question, the authors modify the database schema and create two alternative interpretations. We evaluate on Column, Join and Table ambiguity, with 2948 examples in total. 
    \item  \textit{Ambrosia}~\cite{ambrosia}. Ambrosia is an LLM-generated benchmark with three ambiguity types: scope, attachment, and vagueness. Scope and attachment arise from the natural-language question, whereas vagueness corresponds to schema ambiguity and is most comparable to AmbiQT. We evaluate on the 373 vagueness examples. Some examples include a third interpretation formed by combining the two alternatives, e.g., \texttt{SELECT a, b} in addition to \texttt{SELECT a} and \texttt{SELECT b}. We remove this combined interpretation because it is not independent of the first two and to keep the evaluation consistent with AmbiQT.
\squishend 

\paragraph{Evaluation metric} 
For both benchmarks, we measure how well the prediction set covers the possible interpretations. Following AmbiQT, we use EitherInTopK and BothInTopK, which report the percentage of examples for which at least one or both interpretations appear among the top-K predictions. When a method produces more than K predictions, we evaluate only the first K. Following \cite{logicalbeam}, we set K=5 by default. We also report inference latency, which is measured by comparing each method to vanilla beam search with the same base model.

\paragraph{Baselines and comparison methods} 
We list a series of model-agnostic baselines and comparison methods. Apart from methods that aim to generate multiple interpretations under ambiguity, we also look into state-of-the-art text-to-SQL systems that generate diverse SQL queries. All experiments are conducted on two NVIDIA RTX A6000 GPUs in a Linux environment. We categorize all methods into three classes based on the key techniques they adopt:
\begin{itemize}
    \item \textit{Decoding methods}: Following previous work\cite{logicalbeam}, we adopt beam search and random sampling as two baselines. A beam width of 10 and a temperature of 1.0 are used. We also evaluate model ensemble which is widely adopted in many systems\cite{XiYanSQL, Agentar}. We combine predictions from models with beam search in a round-robin manner after deduplication.
    \item \textit{Fine-tuning-based methods}: These methods rely on task-specific fine-tuned models, including LogicalBeam\cite{logicalbeam} and DFPL\cite{DFPL}.
    \item \textit{Prompt-based methods}: Some methods manipulate the model input to generate different interpretations, usually dependent on carefully designed prompts and exploiting the reasoning ability of LLMs, including ODIN\cite{odin}, CLEAR\cite{clear}. Moreover, we adopt the prompt strategy used in Ambrosia that acknowledges the existence of ambiguity, denoted as Prompt. Some systems also incorporate generators with different reasoning strategies\cite{chase_sql}. We evaluate Deepeye-SQL\cite{deepeye} which adopts skeleton-based, ICL, and divide-and-conquer as three strategies. For each strategy, we sample K times and select candidates in the same way in model ensemble. 
\end{itemize}

\paragraph{Our methods} 
We consider two variants of our method. The first is \textit{Constrained beam search}, which applies SLA to vanilla beam search with \(\alpha=0\), thereby reducing decoding to schema-element constraints without similarity-based logit adjustment. The second is \textit{SchemaBeam} (Section~\ref{sec:app}), which uses all-MiniLM-L6-v2 from Hugging Face to compute similarities between schema elements.

\begin{table*}[t]
\centering
\small
\begin{tabular}{llllllllll}
\toprule
 & & \multicolumn{2}{c}{Column} & \multicolumn{2}{c}{Join} & \multicolumn{2}{c}{Table}& \multicolumn{2}{c}{Vague}\\
\cmidrule(lr){3-4} \cmidrule(lr){5-6}\cmidrule(lr){7-8}\cmidrule(lr){9-10}
 Model  & Method & Either & Both & Either & Both& Either & Both& Either & Both\\ \midrule
 \multirow{4}{*}{Llama-8B} & Sampling & 59.17  & 11.84 & 76.89&23.45 &66.22&17.15&64.99&8.81\\
 & Beam &  57.29 & 10.56 &77.59 & 16.55&65.49&17.15&57.91&8.86\\
  & Constrained & 61.48 & 17.89 &75.52&31.38&\textbf{70.43}&24.77&58.97&8.97\\
& SchemaBeam & \textbf{62.45} & \textbf{38.36} &\textbf{78.28}&\textbf{54.83}&70.22&\textbf{39.03}&\textbf{67.72}&\textbf{23.63}\\ 
 \midrule
 \multirow{4}{*}{QwenCoder-14B}    & Sampling & 63.58  & 14.37 & 90.57&23.33 &74.70&23.79&64.29&6.76\\
 & Beam &  62.93 & 16.36 &87.93 & 40.34&73.18&23.8&60.98&6.32\\
& Constrained & 67.45 & 21.19 &87.59&41.72&74.24&30.20&62.15&6.08\\
 & SchemaBeam & \textbf{68.57} & \textbf{44.48} & \textbf{90.69}& \textbf{76.55}&\textbf{77.21}&\textbf{54.48}&\textbf{67.13}&\textbf{19.22}\\
  \midrule
\multirow{4}{*}{Qwen-27B} & Sampling & 68.30  & 20.71 &90.45 &21.60 &76.23&22.35&67.76&8.83\\
& Beam & 70.59 & 32.39 &90.00&58.62&\textbf{77.63}&29.50&66.85&13.81\\
  & Constrained & 69.46 & 34.97 &90.00&61.03&77.06&33.17&66.67&15.80\\
& SchemaBeam & \textbf{71.23} & \textbf{47.38} &\textbf{92.76}&\textbf{81.72}&75.37&\textbf{53.07}&\textbf{69.04}&\textbf{25.75}\\
 \bottomrule
\end{tabular}
\caption{Top-5 results of different decoding methods.}
\label{tab:main_results}

\end{table*}

\subsection{Performance with different models}

We evaluate decoding methods on three base models: Llama3.1-8B, Qwen2.5-Coder-14B, and Qwen3.6-27B. We use \(\alpha=0.7\) for Qwen2.5-Coder-14B and \(\alpha=0.8\) for the others. Table~\ref{tab:main_results} reports the results; where SchemaBeam improves BothInTopK in every setting; sampling is averaged over three runs.

By constraining models to adhere to database schema elements, both metrics are improved in most cases, especially for BothInTopK. This is because we prevent hallucinated tokens, and simultaneously reduce the search space. SchemaBeam further enhances the performance even for EitherInTopK, which is partly due to cases where the top-1 prediction is wrong but semantically similar to one of the ground truth schema elements. The increased probability of the correct schema elements contributes to the improvement. Other factors such as deduplication, unhelpful token-masking are analyzed in the ablation study. 

However, for stronger base models which are more likely to generate the correct schema element and less likely to hallucinate, the additional diversity in schema element generation does not necessarily boost EitherInTopK. Another reason is that for the remaining incorrect predictions, the errors usually arise from incorrect SQL structures, such as redundant or missing subqueries and clauses, rather than incorrect schema elements. This also explains why the 14B and 27B models with SchemaBeam show similar results. But the interpretation coverage is consistently improved, demonstrating the applicability of our method to different base models, and the advantage over conventional decoding methods.

\subsection{Comparison with other methods}

We further compare SchemaBeam with existing methods for improving interpretation coverage. Prior work lacks a fair, comprehensive comparison because methods often use different base models, metrics, and dataset splits. To control for these factors, we reimplement all open-source methods with Qwen2.5-Coder-14B, which balances effectiveness and efficiency, and evaluate them under the same pipeline. Methods requiring task-specific fine-tuning are evaluated with their released fine-tuned models. For ODIN, whose code is unavailable, we report results with Claude 3 Haiku. Table~\ref{tab:methods_comparison} presents the results; our main findings are as follows:

\textbf{Prompt-based} methods usually incur high latency, which is mainly caused by the increased context length, especially instructions and few-shot examples, and extra reasoning content they generate during the decoding process. However, they do not necessarily achieve superior performance. One important observation is that ensembling generators with different reasoning strategies fails to handle potential schema ambiguity. Moreover, these methods are highly dependent on the model capability, as we find a large performance drop of CLEAR when switching from GPT-4o which is used in the original paper to other weaker base models.

\textbf{Fine-tuning-based} methods fall short in the generalization ability. Logicalbeam which is fine-tuned on SPIDER, fails to generalize to the Ambrosia dataset due to database domain shift and the limitations of the base model. Although DFPL performs well in several datasets, it is incapable of capturing JOIN ambiguity.

\textbf{Decoding} methods are generally more efficient, and show stable performance across diverse datasets. We notice that combining predictions from Llama and QwenCoder yields better results. The improved diversity could be attributed to different training corpora, optimizing objectives, etc.

\begin{table*}[t]
\centering
\begin{small}
\setlength{\tabcolsep}{4pt}
\begin{tabular}{llrcccccccc}
\toprule
 && &  \multicolumn{2}{c}{Column} & \multicolumn{2}{c}{Join} & \multicolumn{2}{c}{Table} & \multicolumn{2}{c}{Vague}\\
\cmidrule(lr){4-5} \cmidrule(lr){6-7} \cmidrule(lr){8-9} \cmidrule(lr){10-11}
 Category&Method  & latency & Either & Both & Either & Both & Either & Both & Either & Both \\
\midrule
\multirow{4}{*}{Prompt-based} &Prompt   & 1.59\(\times\) & 59.87 & 23.77 & 74.88 & 33.45& 71.91&\underline{58.36} &53.95&13.08 \\
&Reasoning ensemble  & 10.9\(\times\) & 65.75 & 15.95 &82.41 & 19.66&76.99&21.74&63.61&6.01\\
&CLEAR  & 6.36\(\times\) & 62.85 & 32.39 & 86.55& 38.62&70.15&36.34&51.2&11.11\\
&ODIN  & - & \underline{67.44} & \underline{43.69} & 87.85&58.33 &73.96&\textbf{59.99}&-&-\\ \midrule

\multirow{2}{*}{Fine-tuning-based}&DFPL  & 2.54\(\times\) & 64.40 & 38.84 & 75.61&6.10 &71.92&36.66&60.12&\textbf{19.64}\\
&LogicalBeam  & 3.96\(\times\) & 65.56 & 27.42 &87.15 & \underline{63.19}&\underline{77.19}&49.75&18.31&2.97\\ \midrule

\multirow{3}{*}{Decoding-based}&Beam   & 1\(\times\) & 62.93 & 16.36 &\underline{87.93} & 40.34&73.18&23.8&60.98&6.32\\
&Model ensemble  & - & 64.06 & 14.99 & 86.90& 48.28&75.72&28.23&\textbf{68.30}&9.48\\
&SchemaBeam  & 1.1\(\times\) & \textbf{68.57} & \textbf{44.48} & \textbf{90.69}& \textbf{76.55}&\textbf{77.21}&54.48&\underline{67.13}&\underline{19.22}\\

\bottomrule
\end{tabular}
\caption{Performance comparison across different methods.}
\label{tab:methods_comparison}

\end{small}
\end{table*}


\subsection{Ablation study}

We conduct an ablation study on Column (C) and Table (T) ambiguity with Llama to examine the impact of SchemaBeam's hyperparameters and components. Similar trends are observed across other datasets and base models.

\begin{figure}[h]
    \centering
    \includegraphics[width=0.9\linewidth]{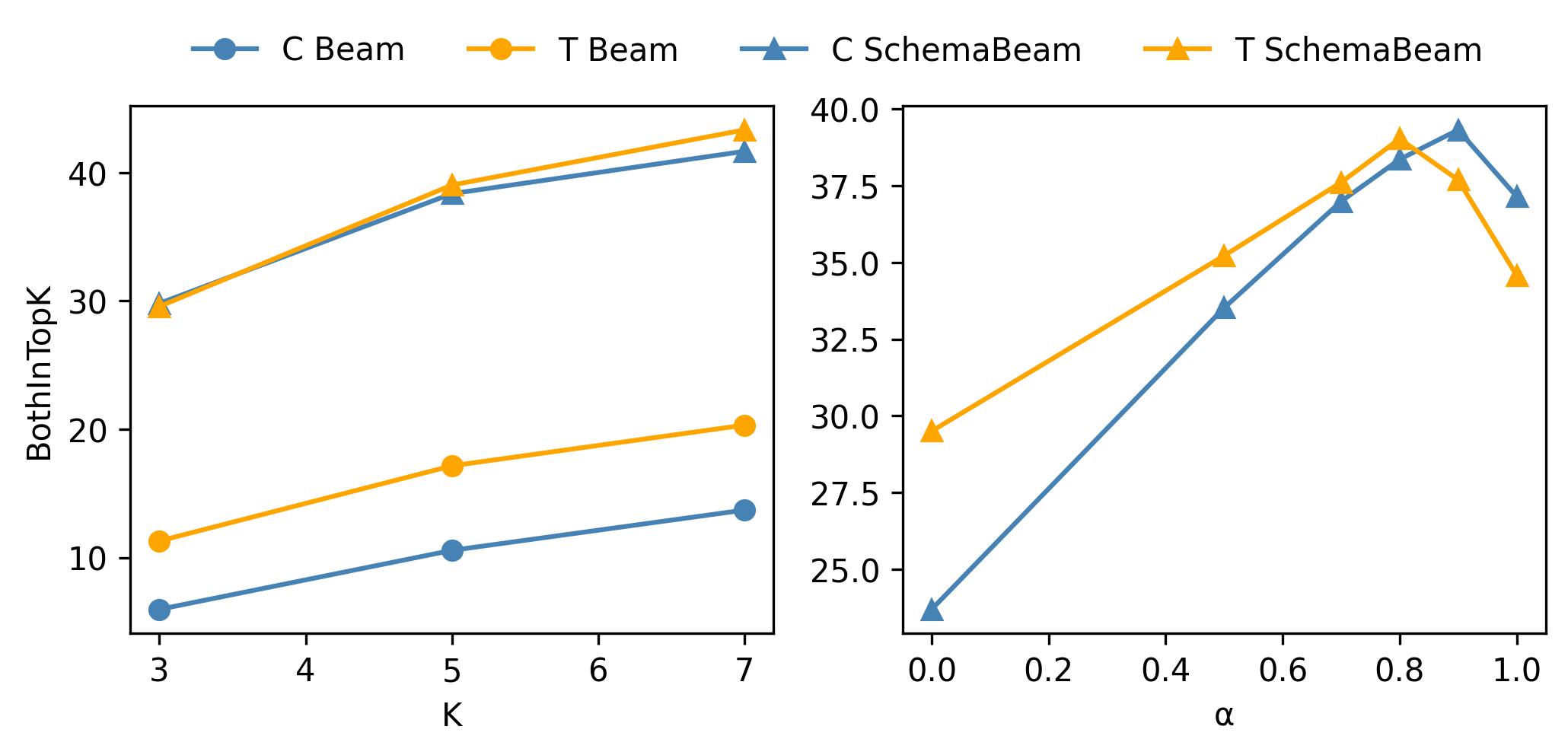}
    \caption{Interpretation coverage vs K and \(\alpha\).}
    \label{fig:topk_C_T}
\end{figure}

\paragraph{Influence of hyper-parameters}
We vary the number of final outputs among 3,5,7 for SchemaBeam and vanilla beam search to analyze where correct predictions are ranked. As shown in Fig. \ref{fig:topk_C_T}(left), the interpretation coverage grows as K increases. But for SchemaBeam, the curve exhibits a steeper slope at the beginning, which indicates the correct interpretations are more likely to be ranked at higher positions. We also report the results with \(\alpha \in [0,1]\) in Fig. \ref{fig:topk_C_T}(right), where we observe a steady growth initially, peaking around \(\alpha=0.8\), and then starts to drop. The main reason is that a large \(\alpha\) distorts the original distribution, which reflects how well each schema element fits the current context.

\paragraph{Effects of each component}
\begin{table}[h]
\centering
\small
\setlength{\tabcolsep}{4pt}
\begin{tabular}{llllll}
\toprule
  &&\multicolumn{2}{c}{Column}  & \multicolumn{2}{c}{Table} \\
\cmidrule(lr){3-4} \cmidrule(lr){5-6}
  Method &Emb.&Either & Both & Either &Both\\ \midrule
  SchemaBeam &Qwen& 62.13 & 38.11& 70.36&38.11\\
  SchemaBeam &mpnet& 62.05 & 39.00& 69.87&38.53\\
  SchemaBeam &MiniLM &62.45 & 38.36& 70.22&39.03\\
  w/ avg&& 62.37 & 38.28& 70.92 &33.87\\
  w/o deduplication&&61.89 &36.18&68.67&35.64\\
  w/o masking&& 61.80&35.78&68.31&34.56\\
  w/o SLA(=beam)&& 57.29&10.56&65.49&17.15\\
 \bottomrule
\end{tabular}
\caption{Component Analysis.}
\label{tab:component_analysis}
\end{table}

Next we analyze each component in SchemaBeam, including the choice of embedding models (all-mpnet-base-v2, Qwen3-Embedding-0.6B), similarity aggregation strategies. We also gradually remove (1) syntactic deduplication, (2) unhelpful token masking and (3) SLA, with the final variant reducing to vanilla beam search. Results in Table \ref{tab:component_analysis} show that our method is not sensitive to the choice of embedding models, as our method relies on the coarse-grained similarity structure rather than exact similarity values, while different embedding models typically introduce only small shifts in similarity scores after normalization. We witness a performance drop in table ambiguity when replacing our similarity aggregation strategy with average pooling, since it weakens the similarity estimation between tables. Finally, all additional designs in SchemaBeam contribute to performance improvements, with SLA providing the largest gain.

\subsection{Case study}
\begin{figure}[h]
    \centering
    \includegraphics[width=1\linewidth]{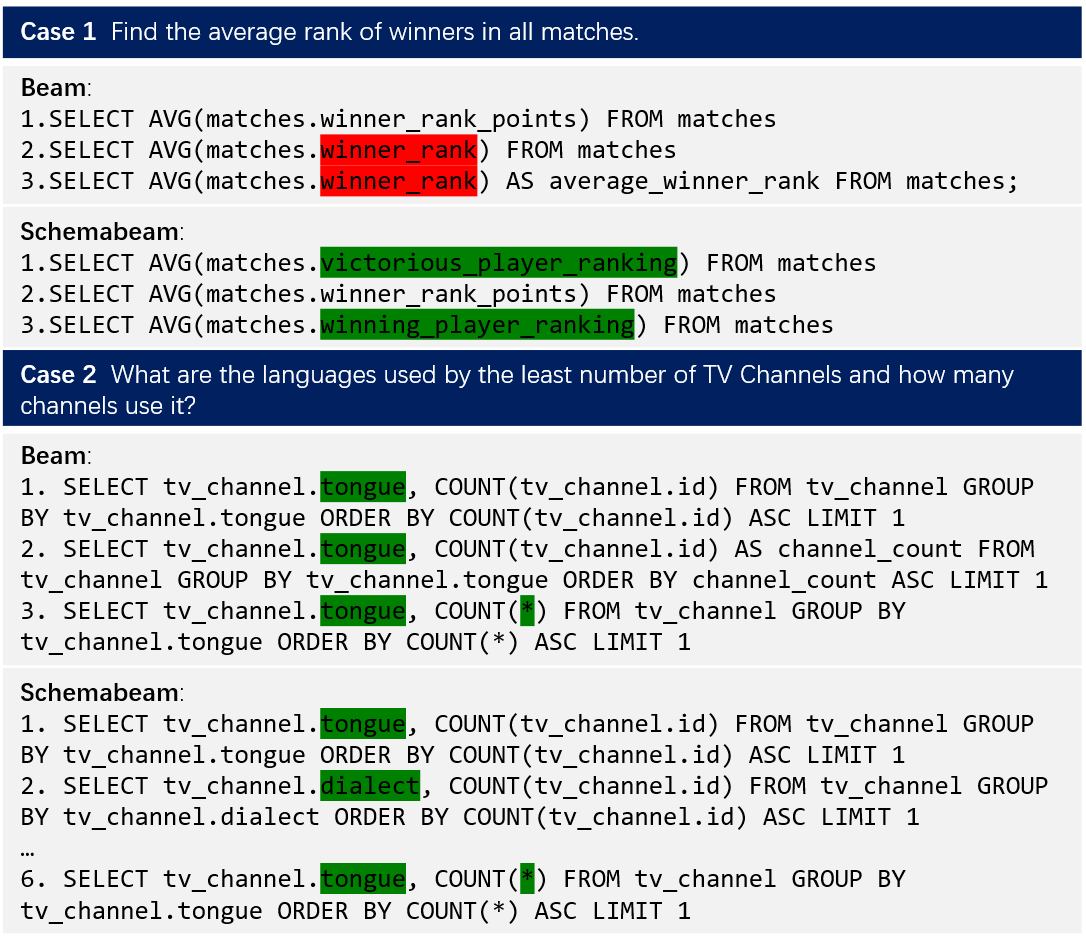}
    \caption{Two examples with vanilla beam search and SchemaBeam. We highlight hallucinated (red) and correct (green) schema elements.}
    \label{fig:case}
\end{figure}

Fig.~\ref{fig:case} shows two representative examples: Case 1, where vanilla beam search fails and SchemaBeam succeeds, and Case 2, where SchemaBeam fails while beam search succeeds. In Case 1, the base model hallucinates columns and selects an incorrect column, whereas SchemaBeam covers both valid interpretations. In Case 2, the correct query requires \texttt{COUNT(*)}, but SchemaBeam ranks it lower because increasing schema-level diversity also introduces randomness in other schema choices. As a result, SchemaBeam detects the ambiguous columns but fails to rank the correct interpretations sufficiently high.

\subsection{Key takeaways}
SchemaBeam consistently improves interpretation coverage. Across three base models and both benchmarks, it improves BothInTopK over vanilla beam search in every setting, often by large margins, and improves EitherInTopK in most cases. The gains are largest for weaker models but remain visible for stronger ones, showing that skewed decoding distributions, rather than model capability alone, limit coverage.

SchemaBeam also provides a strong accuracy--efficiency trade-off. Prompt-based methods are 1.59--10.9\(\times\) slower than beam search and still underperform, while reasoning ensembles do not directly address schema ambiguity. Fine-tuning methods are competitive but brittle: LogicalBeam does not transfer to Ambrosia, and DFPL collapses on JOIN ambiguity. SchemaBeam matches or outperforms these methods at only 1.1\(\times\) beam-search latency, requires no training, and remains robust across datasets.

\textit{Overall}, SchemaBeam shows that lightweight decoding-time intervention can substantially improve ambiguity coverage in text-to-SQL without training, prompting overhead, or dataset-specific failure modes.

\section{Conclusion}
This paper proposes an efficient, training-free decoding method that encourages the exploration of alternative schema interpretations. By leveraging semantic similarities between schema elements through logit adjustment, our method substantially improves interpretation coverage over conventional decoding methods. It achieves performance comparable to fine-tuning and reasoning-based approaches while incurring much lower latency, demonstrating that decoding-time intervention is an effective and practical solution for diverse SQL query generation.


\bibliography{aaai2027}

\begin{thebibliography}{20}
\providecommand{\natexlab}[1]{#1}

\bibitem[{Bhaskar et~al.(2023)Bhaskar, Tomar, Sathe, and Sarawagi}]{logicalbeam}
Bhaskar, A.; Tomar, T.; Sathe, A.; and Sarawagi, S. 2023.
\newblock Benchmarking and improving text-to-sql generation under ambiguity.
\newblock \emph{arXiv preprint arXiv:2310.13659}.

\bibitem[{Chan, Sevastjanova, and El-Assady(2026)}]{pleasqlarify}
Chan, R. S.~M.; Sevastjanova, R.; and El-Assady, M. 2026.
\newblock PleaSQLarify: Visual Pragmatic Repair for Natural Language Database Querying.
\newblock \emph{arXiv preprint arXiv:2603.01795}.

\bibitem[{Ding, Lin, and Zeng(2025)}]{ambisql}
Ding, Z.; Lin, Y.; and Zeng, T. 2025.
\newblock AmbiSQL: Interactive Ambiguity Detection and Resolution for Text-to-SQL.
\newblock \emph{arXiv preprint arXiv:2508.15276}.

\bibitem[{Floratou et~al.(2024)Floratou, Psallidas, Zhao, Deep, Hagleither, Tan, Cahoon, Alotaibi, Henkel, Singla et~al.}]{floratou2024nl2sql}
Floratou, A.; Psallidas, F.; Zhao, F.; Deep, S.; Hagleither, G.; Tan, W.; Cahoon, J.; Alotaibi, R.; Henkel, J.; Singla, A.; et~al. 2024.
\newblock Nl2sql is a solved problem... not!
\newblock In \emph{CIDR}.

\bibitem[{Khattab and Zaharia(2020)}]{colbert}
Khattab, O.; and Zaharia, M. 2020.
\newblock Colbert: Efficient and effective passage search via contextualized late interaction over bert.
\newblock In \emph{Proceedings of the 43rd International ACM SIGIR conference on research and development in Information Retrieval}, 39--48.

\bibitem[{Li et~al.(2026)Li, Chen, Xue, Mei, and Luo}]{deepeye}
Li, B.; Chen, C.; Xue, Z.; Mei, Y.; and Luo, Y. 2026.
\newblock {DeepEye-SQL:} A Software-Engineering-Inspired Text-to-SQL Framework.
\newblock \emph{Proc. {ACM} Manag. Data}, 4(3).

\bibitem[{Li et~al.(2023)Li, Hui, Qu, Yang, Li, Li, Wang, Qin, Geng, Huo et~al.}]{bird}
Li, J.; Hui, B.; Qu, G.; Yang, J.; Li, B.; Li, B.; Wang, B.; Qin, B.; Geng, R.; Huo, N.; et~al. 2023.
\newblock Can llm already serve as a database interface? a big bench for large-scale database grounded text-to-sqls.
\newblock \emph{Advances in Neural Information Processing Systems}, 36: 42330--42357.

\bibitem[{Liu et~al.(2026)Liu, Zhu, Gao, Luo, Li, Shi, Hong, Gao, Li, Ding, and Zhou}]{XiYanSQL}
Liu, Y.; Zhu, Y.; Gao, Y.; Luo, Z.; Li, X.; Shi, X.; Hong, Y.; Gao, J.; Li, Y.; Ding, B.; and Zhou, J. 2026.
\newblock XiYan-SQL: A Novel Multi-Generator Framework for Text-to-SQL.
\newblock \emph{IEEE Transactions on Knowledge and Data Engineering}, 1--14.

\bibitem[{Pourreza et~al.(2024)Pourreza, Li, Sun, Chung, Talaei, Kakkar, Gan, Saberi, Ozcan, and Arik}]{chase_sql}
Pourreza, M.; Li, H.; Sun, R.; Chung, Y.; Talaei, S.; Kakkar, G.~T.; Gan, Y.; Saberi, A.; Ozcan, F.; and Arik, S.~O. 2024.
\newblock Chase-sql: Multi-path reasoning and preference optimized candidate selection in text-to-sql.
\newblock \emph{arXiv preprint arXiv:2410.01943}.

\bibitem[{Qiu et~al.(2025)Qiu, Li, Su, and Chen}]{EIG_ambi}
Qiu, L.; Li, J.; Su, C.; and Chen, L. 2025.
\newblock Interactive Text-to-SQL via Expected Information Gain for Disambiguation.
\newblock \emph{arXiv preprint arXiv:2507.06467}.

\bibitem[{Saparina and Lapata(2024)}]{ambrosia}
Saparina, I.; and Lapata, M. 2024.
\newblock Ambrosia: A benchmark for parsing ambiguous questions into database queries.
\newblock \emph{Advances in Neural Information Processing Systems}, 37: 90600--90628.

\bibitem[{Saparina and Lapata(2025{\natexlab{a}})}]{DFPL}
Saparina, I.; and Lapata, M. 2025{\natexlab{a}}.
\newblock Disambiguate first, parse later: Generating interpretations for ambiguity resolution in semantic parsing.
\newblock In \emph{Findings of the Association for Computational Linguistics: ACL 2025}, 16825--16839.

\bibitem[{Saparina and Lapata(2025{\natexlab{b}})}]{IntentRL}
Saparina, I.; and Lapata, M. 2025{\natexlab{b}}.
\newblock Reasoning About Intent for Ambiguous Requests.
\newblock \emph{arXiv preprint arXiv:2511.10453}.

\bibitem[{Sharma et~al.(2025)Sharma, Narayanam, Pal, Yeturu, Saini, and Mukherjee}]{TTD-SQL}
Sharma, C.; Narayanam, R.; Pal, S.; Yeturu, K.; Saini, S.~K.; and Mukherjee, K. 2025.
\newblock TTD-SQL: Tree-Guided Token Decoding for Efficient and Schema-Aware SQL Generation.
\newblock In \emph{Proceedings of the 2025 Conference on Empirical Methods in Natural Language Processing: Industry Track}, 1287--1298.

\bibitem[{Vaidya et~al.(2025)Vaidya, Sankararaman, Ding, Lei, Qin, Narayanaswamy, and Kraska}]{odin}
Vaidya, K.; Sankararaman, A.; Ding, J.; Lei, C.; Qin, X.; Narayanaswamy, B.; and Kraska, T. 2025.
\newblock ODIN: A NL2SQL Recommender to Handle Schema Ambiguity.
\newblock \emph{arXiv preprint arXiv:2505.19302}.

\bibitem[{Wang et~al.(2023)Wang, Gao, Li, and Lou}]{know_what_idk}
Wang, B.; Gao, Y.; Li, Z.; and Lou, J.-G. 2023.
\newblock Know what I don’t know: Handling ambiguous and unknown questions for text-to-SQL.
\newblock In \emph{Findings of the Association for Computational Linguistics: ACL 2023}, 5701--5714.

\bibitem[{Wang et~al.(2025)Wang, Sun, Dong, Dai, Yuan, Chu, Gao, Qi, Zhang, and Yan}]{Agentar}
Wang, P.; Sun, B.; Dong, X.; Dai, Y.; Yuan, H.; Chu, M.; Gao, Y.; Qi, X.; Zhang, P.; and Yan, Y. 2025.
\newblock Agentar-Scale-SQL: Advancing Text-to-SQL through Orchestrated Test-Time Scaling.
\newblock arXiv:2509.24403.

\bibitem[{Yu et~al.(2018)Yu, Zhang, Yang, Yasunaga, Wang, Li, Ma, Li, Yao, Roman et~al.}]{spider}
Yu, T.; Zhang, R.; Yang, K.; Yasunaga, M.; Wang, D.; Li, Z.; Ma, J.; Li, I.; Yao, Q.; Roman, S.; et~al. 2018.
\newblock Spider: A large-scale human-labeled dataset for complex and cross-domain semantic parsing and text-to-sql task.
\newblock In \emph{Proceedings of the 2018 conference on empirical methods in natural language processing}, 3911--3921.

\bibitem[{Zhang et~al.(2025)Zhang, Ma, Xu, Zhang, Peng, and Jin}]{clear}
Zhang, M.; Ma, K.; Xu, L.; Zhang, K.; Peng, Y.; and Jin, R. 2025.
\newblock CLEAR: A Parser-Independent Disambiguation Framework for NL2SQL.
\newblock In \emph{2025 IEEE 41st International Conference on Data Engineering (ICDE)}, 1--14. IEEE.

\bibitem[{Zhang et~al.(2019)Zhang, Kishore, Wu, Weinberger, and Artzi}]{bertscore}
Zhang, T.; Kishore, V.; Wu, F.; Weinberger, K.~Q.; and Artzi, Y. 2019.
\newblock Bertscore: Evaluating text generation with bert.
\newblock \emph{arXiv preprint arXiv:1904.09675}.

\end{thebibliography}


\end{document}